\documentclass[11pt,a4paper]{article}
\usepackage[hyperref]{acl2020}

 \usepackage{CJKutf8}
\usepackage{graphicx}
\usepackage{epstopdf}
\usepackage{times}
\usepackage{url}
\usepackage{latexsym}
\usepackage{amsfonts}
\usepackage{xspace}
\usepackage{booktabs}
\usepackage{color}
\usepackage{graphicx}
\usepackage{tabulary}
\usepackage{enumitem}
\usepackage{url}
\usepackage{tikz} 
\usepackage{amsfonts}

\newenvironment{tightitemize}%
  {\begin{itemize}[topsep=0pt, partopsep=0pt] %
    \setlength{\itemsep}{0pt}%
    \setlength{\parskip}{0pt}%
    }%
  {\end{itemize}}
  \usepackage{booktabs}
\usepackage{subcaption}
\usepackage{lipsum}

\usepackage{parskip}
  \usepackage{color}
  {\begin{enumerate}â \footnotesize%
    \setlength{\itemsep}{0pt}%
    \setlength{\parskip}{0pt}%
    }%
  {\end{enumerate}}


\usepackage{microtype}
\usepackage{multirow}
\usepackage{verbatim}
\usepackage{amsmath,amsthm,amssymb}
\usepackage{array}
\usepackage[scaled=0.86]{helvet}
\usepackage{ifthen}
\usepackage{courier}
\usepackage[boxed]{algorithm}
\usepackage{caption}
\usepackage{algpseudocode}
\belowdisplayskip \abovedisplayskip

\usepackage{amsmath}

\aclfinalcopy % Uncomment this line for the final submission
\def\aclpaperid{72} %  Enter the acl Paper ID here

\title{Dice Loss for Data-imbalanced NLP Tasks}
\author{
Xiaoya Li$^{\clubsuit}$,
Xiaofei Sun$^{\clubsuit}$,
Yuxian Meng$^{\clubsuit}$, {Guoyin Wang}$^{\blacklozenge}$ \\
{\bf {Junjun Liang}$^{\clubsuit}$,
{ Fei Wu}$^{\spadesuit}$ and { Jiwei Li}$^{\spadesuit\clubsuit}$}  \\
$^{\spadesuit}$ Department of Computer Science and Technology, Zhejiang University\\
$^{\clubsuit}$ Shannon.AI, $^\blacklozenge$Amazon  ~~\\
 {\{xiaoya\_li, xiaofei\_sun, yuxian\_meng,  jiwei\_li\}}@shannonai.com, 
 wufei@cs.zju.edu.cn 
}

\begin{document}
\maketitle
\begin{abstract}
Many NLP tasks such as 
tagging and machine reading comprehension (MRC) 
are faced with the severe  data imbalance issue:  negative examples significantly outnumber positive ones, and the huge number of 
easy-negative examples overwhelms training. 
The most commonly used cross entropy criteria 
 is actually accuracy-oriented, which creates a discrepancy between training and test. At training time, each training instance contributes equally to the objective function, while at test time F1 score concerns more about positive examples. 

In this paper, we propose 
to use dice loss
in replacement of the standard cross-entropy objective for data-imbalanced NLP tasks. 
Dice loss
is  based on the
Sørensen–Dice coefficient \cite{sorensen1948method} or Tversky index \cite{tversky1977features}, which
attaches similar importance to  
false positives and false negatives, 
and is  more immune to the 
data-imbalance issue. 
To further
alleviate the dominating influence from easy-negative examples in training, 
we propose to associate training examples with dynamically adjusted weights to deemphasize easy-negative examples.
Experimental results show that this strategy narrows down the gap between the F1  score in evaluation and the dice  loss in training.

With the proposed training objective, we  
 observe  significant performance boosts over a wide range of data imbalanced  NLP tasks. 
 Notably, we are able to achieve  SOTA results 
 on CTB5,  CTB6 and UD1.4 for the part of speech tagging  task, and  
  competitive or even better results on CoNLL03, OntoNotes5.0, MSRA and 
OntoNotes4.0 for the 
named entity recognition  task along with the machine reading comprehension and paraphrase identification tasks. The code can be found at \url{https://github.com/ShannonAI/dice_loss_for_NLP}.
\end{abstract}

\section{Introduction}

Data imbalance  is a common issue in a variety of NLP tasks such as 
  tagging and machine reading comprehension.
 Table \ref{num-stat}  gives concrete examples: 
for the Named Entity Recognition (NER) task \cite{sang2003introduction,nadeau2007survey}, most tokens are backgrounds  with tagging class $O$.
Specifically, 
the number of tokens with tagging class $O$  is 5 times as many as those with 
entity
 labels for the CoNLL03 dataset and 8 times for the OntoNotes5.0 dataset; 
Data-imbalanced issue is  more severe for
 MRC tasks \cite{rajpurkar2016squad,nguyen2016ms,rajpurkar2018know,kovcisky2018narrativeqa,dasigi2019quoref} with the value of negative-positive ratio being 50-200, which is due to the reason that
the task of MRC is usually formalized as predicting the {\it starting} and {\it ending} indexes  conditioned on the query and the context, and given a chunk of text of an arbitrary length, only two tokens are positive (or of interest) with all the rest 
being background. 

\begin{table}
\center
\resizebox{\linewidth}{10mm}{
\begin{tabular}{|c|c|c|c|}\hline
{\bf Task} & {\bf $\#$ neg} & {\bf $\#$ pos } & {\bf ratio}\\\hline
CoNLL03 NER&170K & 34K&4.98 \\\hline
OntoNotes5.0 NER&1.96M& 239K& 8.18 \\\hline
SQuAD 1.1  \cite{rajpurkar2016squad}&10.3M& 175K&  55.9 \\\hline
SQuAD 2.0  \cite{rajpurkar2018know}&15.4M&188K&82.0\\\hline
QUOREF \cite{dasigi2019quoref} &6.52M & 38.6K&169\\\hline 
\end{tabular}
}
\caption{Number of positive and negative examples and their ratios for different data-imbalanced NLP tasks. }
\label{num-stat}
\vskip -0.15in
\end{table}

Data imbalance results in 
the following 
two   issues:
(1) {\bf the training-test discrepancy}: 
Without balancing the labels,
the learning process tends to converge to a point that strongly biases towards class with the majority label.  
This actually creates a discrepancy 
between training and test: at training time, each training instance contributes equally to the objective function, whereas at test time, F1 gives equal weight to positive and negative examples; (2) {\bf the overwhelming effect of easy-negative examples}. 
As pointed out by \newcite{meng2019dsreg}, a significantly large number of negative examples also means that
the number of easy-negative example is large. 
The huge number of easy examples tends to overwhelm the training, making the model not sufficiently learn to  distinguish between positive examples and hard-negative examples. 
The cross-entropy objective (CE for short) or maximum likelihood (MLE) objective, 
which is widely adopted as the training objective 
 for data-imbalanced NLP tasks \cite{lample2016neural,wu2019glyce,devlin2018bert,yu2018qanet,mccann2018natural,ma2016end,chen2017reading},  handles neither of the issues.

To handle the first issue, 
we propose to replace CE or MLE with  
losses  based on the
 Sørensen–Dice coefficient \cite{sorensen1948method} or Tversky index \cite{tversky1977features}. 
The Sørensen–Dice coefficient, dice loss for short, is the harmonic mean of precision and recall.
It attaches 
equal importance to  false positives (FPs) and false negatives (FNs) and is thus  more immune to 
 data-imbalanced datasets. 
Tversky index extends dice loss by using a weight that trades precision and recall, which can be thought as the approximation of the $F_{\beta}$ score, and thus comes with more flexibility. 
Therefore, we use dice loss
or Tversky index
 to replace CE loss to address the first issue.

Only using dice loss
or Tversky index is not enough since 
they are unable to address the dominating influence of easy-negative examples.
This is
intrinsically 
 because dice loss is actually 
a soft version of the F1 score. Taking the binary classification task as an example, at test time, an example will be classified as negative as long as its probability is smaller than 0.5, 
but training will push the value to 0 as much as possible. 
This gap isn't a big issue for balanced datasets, but is extremely detrimental if a big proportion of training examples are easy-negative ones: 
easy-negative examples can easily dominate training 
since their probabilities can be pushed to 0 fairly easily.
Meanwhile, 
the model can hardly distinguish between 
hard-negative examples and positive ones. 
Inspired by the idea of focal loss \cite{lin2017focal} in computer vision, we propose
a dynamic weight adjusting strategy, which
associates each training example with a 
 weight in proportion to $(1-p)$, and this weight dynamically changes as training proceeds. 
  This strategy helps deemphasize confident examples during training as their probability $p$ approaches  $1$,  making the model attentive to hard-negative examples, and thus alleviates the dominating effect of easy-negative  examples. Combing both strategies, we observe significant performance boosts on a wide range of data imbalanced NLP tasks.

% Notably, we are able to achieve  SOTA results 
%  on CTB5 (97.92, +1.86), CTB6 (96.57, +1.80) and UD1.4 (96.98, +2.19) for the 
%   POS  task;  competitive or better results over state-of-the-art models on CoNLL03 (93.33, +0.29), OntoNotes5.0 (92.07, +0.96), MSRA (96.72, +0.97) and 
% OntoNotes4.0 (84.47,+2.36) for the 
% NER  task,  along with competitive results on the tasks of machine reading comprehension and paraphrase identification.
 
The rest of this paper is organized as follows: 
related work is presented in Section 2. 
We describe different proposed losses in Section 3. Experimental results are presented in Section 4. We perform ablation studies in Section 5,  followed by a brief conclusion in Section 6.

\section{Related Work}
\subsection{Data Resampling}
The idea of weighting  training examples has a long history. Importance sampling \cite{resample1953} assigns weights to different samples and changes the data distribution. 
Boosting algorithms such as AdaBoost \cite{adboost} select harder examples to train subsequent classifiers. Similarly, hard example mining \cite{hard2011example} downsamples the majority class and exploits the most difficult examples.
 Oversampling \cite{Chen2010RAMOBoostRM,Chawla2002SMOTESM} is  used to balance the data distribution. Another line of data resampling is to dynamically control the weights of examples as training proceeds. For example, 
focal loss \cite{lin2017focal} used a soft weighting scheme that emphasizes harder examples during training. In self-paced learning \cite{self2010}, example weights are obtained through optimizing the weighted training loss which encourages learning easier examples first. At each training step, self-paced learning algorithm   optimizes  model parameters and example weights jointly.  
Other works  \cite{Chang2017ActiveBT, Katharopoulos2018NotAS}  adjusted the weights of different training examples based on training loss. Besides, 
recent work \cite{Jiang2017MentorNetLD,Fan2006LearningTT} proposed to learn a separate network to predict sample weights.
\subsection{Data Imbalance Issue in Computer Vision}
The background-object label imbalance issue is severe and thus well studied in the 
field of 
object detection \cite{7299170, Girshick2015FastR, He2015DeepRL,Girshick2013RichFH,Ren2015FasterRT}. 
The idea of hard negative mining (HNM) \cite{Girshick2013RichFH} has gained much attention  recently. 
%%%%%%%%\newcite{shrivastava2016ohem} proposed the online hard example mining (OHEM) algorithm in an iterative manner
%%%%%%%that makes training progressively more difficult, and pushes the model to learn better.
%%%%%%%%\newcite{ssd2016liu}  sorted all of the negative samples based on the  confidence loss  and picked the training examples with  the negative-positive ratio at  3:1.  
\newcite{pang2019rcnn} proposed a novel method called
IoU-balanced sampling and
\newcite{aploss2019chen} designed a ranking model to replace the
conventional classification task with an average-precision
loss to alleviate the  class imbalance issue. 
The efforts made on object detection have greatly inspired us to solve the data imbalance issue in NLP.

\newcite{sudre2017} addressed the severe class imbalance issue for the image segmentation task. They proposed to use the class re-balancing property of the Generalized Dice Loss as the training objective for unbalanced tasks. 
\newcite{shen2018organseg} investigated the influence of Dice-based loss for multi-class organ segmentation using a dataset of abdominal CT volumes. 
\newcite{oldrich2018batch} proposed to use the batch soft Dice loss function to train the CNN network for the task of segmentation of organs at risk (OAR) of medical images. 
\newcite{reuben2019continue} extended the definition of the classical Dice coefficient to facilitate the direct comparison of a ground truth binary image with a probabilistic map. 
In this paper, we introduce dice loss into NLP tasks as the training objective and propose a dynamic weight adjusting strategy to address the dominating influence of easy-negative examples.

\section{Losses}
\subsection{Notation}
For illustration purposes, 
we use the binary classification task to demonstrate how different losses work.
The mechanism can be easily extended to multi-class classification. 
Let $X$ denote a set of training instances and each instance $x_i\in  X$ is associated with a golden binary label $y_i=[y_{i0},y_{i1}]$ denoting the ground-truth class $x_i$ belongs to, and $p_i=[p_{i0},p_{i1}]$ is the predicted probabilities of the two classes respectively, where $y_{i0},y_{i1}\in \{0,1\}, p_{i0},p_{i1}\in[0,1]$ and $p_{i1}+p_{i0}=1$.

\subsection{Cross Entropy Loss}
The vanilla cross entropy (CE) loss is given by:
\begin{equation}
\text{CE} =-\frac{1}{N} \sum_{i}\sum_{j\in\{0,1\}}y_{ij}\log {p}_{ij}
\label{eq1}
\end{equation}
As can be seen from Eq.\ref{eq1}, each $x_i$ contributes equally to the final objective. 
Two strategies are normally used to address the
the case where we wish that not all $x_i$ are treated equally:
associating 
  different classes with different weighting factor $\alpha$ or resampling the datasets.
For the former, Eq.\ref{eq1} is adjusted as follows:
\begin{equation}
\text{Weighted~~CE} = -\frac{1}{N} \sum_{i}\alpha_i\sum_{j\in\{0,1\}}y_{ij}\log {p}_{ij}
\label{eq2}
\end{equation}
where $\alpha_i\in[0,1]$ may be set by the inverse class frequency or treated as a hyperparameter to set by cross validation. In this work, we use $\lg (\frac{n-n_t}{n_t}+K)$ to calculate the coefficient $\alpha$, where $n_t$ is the number of samples with class $t$ and $n$ is the total number of samples in the training set. $K$ is a hyperparameter to tune. Intuitively, this equation assigns less weight to the majority class and more weight to the minority class.
The  data resampling strategy constructs a new dataset by sampling training examples from the original dataset based on human-designed criteria, e.g. 
extracting equal training samples from each class. 
Both  strategies are 
equivalent to changing the data distribution during training
and thus are 
 of the same nature. 
Empirically, these two methods are not widely used due to the trickiness of selecting $\alpha$ especially for multi-class classification tasks and that inappropriate selection can easily bias towards rare classes \cite{valverde2017improving}. 

\subsection{Dice Coefficient and Tversky Index }
\label{sec33}
Sørensen–Dice coefficient \cite{sorensen1948method,dice1945measures}, dice coefficient (DSC) for short, is an 
F1-oriented 
 statistic used to gauge the similarity of two sets. Given two sets $A$ and $B$, the vanilla dice coefficient between them is given as follows:
 \begin{equation}
 \text{DSC}(A,B) = \frac{2|A\cap B|}{|A|+|B|}
 \end{equation}
In our case, $A$ is the set that contains all positive examples predicted by a specific model, and $B$ is the  set of all golden positive  examples in the dataset. 
When applied to boolean data with the definition of true positive (TP), false positive (FP), and false negative (FN), it can be then written as follows:
 \begin{equation}
 \begin{aligned}
 \text{DSC} &   =\frac{2\text{TP}}{2\text{TP}+\text{FN}+\text{FP}} 
 = \frac{2 \frac{\text{TP}}{\text{TP}+\text{FN}} \frac{\text{TP}}{\text{TP}+\text{FP}} }{\frac{\text{TP}}{\text{TP}+\text{FN}}+\frac{\text{TP}}{\text{TP}+\text{FP}}}\\&= \frac{2 \text{Pre} \times \text{Rec}}{\text{Pre+Rec}}= F1
 \end{aligned}
\end{equation}
For an individual example $x_i$, its corresponding dice coefficient is given as follows: 
\begin{equation}
\text{DSC}(x_i) = \frac{2{p}_{i1} y_{i1}}{ {p}_{i1}+ y_{i1}}
\label{dssafc}
\end{equation}
As can be seen, a negative example ($y_{i1}=0$) does not contribute to the objective. 
For smoothing purposes,  it is common to add a $\gamma$ factor to both the nominator and the denominator, making the form to be as follows (we simply set $\gamma=1$ in the rest of this paper): 
\begin{equation}
\text{DSC}(x_i) = \frac{2{p}_{i1} y_{i1}+\gamma}{ {p}_{i1}+ y_{i1}+\gamma}
\end{equation}
As can be seen, negative examples whose  DSC is  $\frac{\gamma}{ {p}_{i1}+\gamma}$, 
 also contribute to the training. 
Additionally, \newcite{milletari2016v} proposed to change the denominator to the square form for faster convergence, which leads
to the following dice loss (DL):
\begin{equation}
\text{DL}=\frac{1}{N} \sum_{i} \left[ 1 -  \frac{2{p}_{i1} y_{i1}+\gamma}{{p}_{i1}^2 + y_{i1}^2+\gamma}\right]
\end{equation}
Another version of DL is to directly compute set-level dice coefficient instead of the sum of individual dice coefficient, which is easier for optimization:
\begin{equation}
    \text{DL} =  1 -  \frac{2 \sum_i p_{i1}  y_{i1}+\gamma}{\sum_ip_{i1}^2  +\sum_iy_{i1}^2+\gamma}
\end{equation}

\begin{table}[t]
\centering
\begin{tabular}{rl}
    \toprule
    {\bf Loss} & {\bf Formula (one sample $x_i$)} \\
    \midrule
    CE & $-\sum_{j\in\{0,1\}}y_{ij}\log {p}_{ij}$\\
    WCE & $-\alpha_i\sum_{j\in\{0,1\}}y_{ij}\log {p}_{ij}$ \\
    DL&$1-\frac{2p_{i1}  y_{i1}+\gamma}{p_{i1}^2  +y_{i1}^2+\gamma}$\\
    TL & $1-\frac{ p_{i1} y_{i1}+\gamma }{p_{i1}y_{i1}+\alpha~p_{i1}y_{i0}+\beta~p_{i0}
    y_{i1}+\gamma} $\\
    DSC & $1- \frac{2(1-p_{i1})^{\alpha} p_{i1}\cdot y_{i1}+\gamma}{(1-p_{i1})^{\alpha} p_{i1}+y_{i1}+\gamma}$\\
    FL & $-\alpha_i\sum_{j\in\{0,1\}}(1-p_{ij})^\gamma\log p_{ij}$\\
    \bottomrule
\end{tabular}
\caption{Different losses and their formulas. We add +1 to DL, TL and DSC so that they are positive.}
\label{tab:losses}
\vskip -0.15in
\end{table}

Tversky index (TI), which can be thought as the approximation of the $F_{\beta}$ score, extends dice coefficient to a more general case.
Given two sets $A$ and $B$, tversky index is computed as follows:
\begin{equation}
 \text{TI} = \frac{|A\cap B|}{|A\cap B|+ \alpha|A\backslash B|+\beta|B\backslash A|}
\end{equation}
Tversky index offers the flexibility in controlling 
the tradeoff between false-negatives and false-positives. It
 degenerates to DSC if $\alpha=\beta=0.5$. 
The Tversky loss (TL) is thus given  as follows:
\begin{equation}
\small
\text{TL} = \frac{1}{N} \sum_{i} \left[ 1 - \frac{ p_{i1} y_{i1}+\gamma }{p_{i1}y_{i1}+\alpha~p_{i1}y_{i0}+ \beta~p_{i0} y_{i1}+\gamma } \right] 
\end{equation}
\subsection{Self-adjusting Dice Loss}
Consider a simple case where the dataset consists of only one example $x_i$, which is classified as positive as long as $p_{i1}$ is larger than 0.5. 
The computation of $F1$ score is actually as follows:
\begin{equation}
\text{F1}(x_i) = 2~\frac{\mathbb{I}( p_{i1}>0.5) y_{i1}}{ \mathbb{I}( p_{i1}>0.5) +y_{i1}}
\label{soft}
\end{equation}
Comparing Eq.\ref{dssafc} with Eq.\ref{soft}, we can see that 
Eq.\ref{dssafc}
 is actually a soft form of $F1$, using
a continuous 
 $p$ rather than the  binary  
$\mathbb{I}( p_{i1}>0.5)$. 
This gap isn't a big issue for balanced datasets, but is extremely detrimental if a big proportion of training examples are easy-negative ones: 
easy-negative examples can easily dominate training 
since their probabilities can be pushed to 0 fairly easily. Meanwhile, 
the model can hardly distinguish between 
hard-negative examples and positive ones, which has a huge negative effect on the final F1 performance.

 \begin{table*}[t]
\footnotesize
\centering
\setlength{\tabcolsep}{4pt}
\resizebox{\linewidth}{28mm}{
\begin{tabular}{l@{\hspace{0.3cm}}c@{\hspace{0.25cm}}c@{\hspace{0.25cm}}c@{\hspace{0.25cm}}c@{\hspace{0.25cm}}c@{\hspace{0.25cm}}c@{\hspace{0.25cm}}c@{\hspace{0.25cm}}c@{\hspace{0.25cm}}c@{\hspace{0.25cm}}}
\toprule
 & \multicolumn{3}{c}{\bf CTB5}& \multicolumn{3}{c}{\bf CTB6}& \multicolumn{3}{c}{\bf UD1.4} \\\hline
%\multicolumn{4}{c}{Chinese CTB5}\\\hline
{\bf Model} & {\bf Prec.} & {\bf Rec.} & {\bf F1} & {\bf Prec.} & {\bf Rec.} & {\bf F1} & {\bf Prec.} & {\bf Rec.} & {\bf F1}  \\
\midrule
Joint-POS(Sig)\cite{shao2017character} &93.68&94.47&94.07 &-&-&90.81 &89.28&89.54&89.41\\
Joint-POS(Ens)\cite{shao2017character} &93.95&94.81&94.38&-&-&-&89.67&89.86&89.75\\
Lattice-LSTM\cite{lattice2018zhang}  & 94.77& 95.51& 95.14& 92.00& 90.86&91.43&90.47&89.70& 90.09 \\
BERT-Tagger\cite{devlin2018bert}& 95.86 & 96.26 & {\bf 96.06} & 94.91 & 94.63 & {\bf 94.77} & 95.42 & 94.17 & {\bf 94.79}\\
\hline
% BERT+WeightCE & 96.45 & 96.41 & 96.43 & 95.34 & 96.22 & 95.78 & 96.09 & 97.08 & 96.58  \\
%  &  &  & (+0.37) &  &  & (+1.01) &  &  & (+1.79)  \\\hline
BERT+FL & 96.11 & 97.42 &  96.76& 95.80 & 95.08 &  95.44 & 96.33 & 95.85 & 96.81 \\
 & & &  (+0.70)&  &  &  (+0.67) &  &  & (+2.02) \\\hline
BERT+DL & 96.77  & 98.87 & 97.81& 94.08 & 96.12 & 95.09  & 96.10 & 97.79 & 96.94 \\
&  &  & (+1.75)&  &  & (+0.32)  &  &  & (+2.15) \\\hline
BERT+DSC & 97.10 &98.75& \textbf{97.92} & 96.29& 96.85& \textbf{96.57}  & 96.24 & 97.73 & \textbf{96.98} \\
 &  && \textbf{(+1.86)} & & & \textbf{(+1.80)}  & &  & \textbf{(+2.19)}
\\
\bottomrule\hline

\end{tabular}
}
\caption{Experimental results for Chinese POS datasets including CTB5, CTB6 and UD1.4.}
\label{tab:pos_result}
\vskip -0.15in
\end{table*} 
 \begin{table}[t]
\footnotesize
\centering
\setlength{\tabcolsep}{4pt}
\begin{tabular}{l@{\hspace{0.25cm}}c@{\hspace{0.25cm}}c@{\hspace{0.25cm}}c}
\toprule
\multicolumn{4}{c}{\bf English WSJ}\\\hline
{\bf Model} & {\bf Prec.} & {\bf Rec.} & {\bf F1}  \\\hline
Meta BiLSTM\cite{pos-wsj}   & -  & -  & 98.23 \\
BERT-Tagger \cite{devlin2018bert} &
99.21 & 98.36 & {\bf 98.86} \\\hline 
BERT-Tagger+FL & 98.36 & 98.97 & 98.88  \\
 &  &  & (+0.02)  \\\hline
BERT-Tagger+DL & 99.34 & 98.22 & 98.91 \\
 &  &  & (+0.05) \\\hline
BERT-Tagger+DSC   & {99.41} & {98.93} & \textbf{99.38}  \\
  &  &  & \textbf{(+0.52)} \\\hline

\multicolumn{4}{c}{\bf English Tweets }\\\hline
{\bf Model} & {\bf Prec.} & {\bf Rec.} & {\bf F1}  \\\hline 
FastText+CNN+CRF\cite{twitter-pos} &  - & - & 91.78 \\
BERT-Tagger \cite{devlin2018bert} & 92.33 &  91.98 & {\bf 92.34} \\\hline
BERT-Tagger+FL & 91.24 &  93.22  & 92.47  \\ 
 &  &  & (+0.13)  \\\hline
BERT-Tagger+DL &  91.44  &  92.88   & 92.52 \\
 &  &  & (+0.18) \\\hline
BERT-Tagger+DSC   & {92.87} & {93.54} & \textbf{92.58}  \\
  &  &  & \textbf{(+0.24)}  \\

\bottomrule
\end{tabular}

\caption{Experimental results for English POS datasets.}
\label{tab:eng_pos_result}
\vskip -0.15in
\end{table} 
\begin{table}[t]
\footnotesize
\centering
\resizebox{\columnwidth}{90mm}{
\setlength{\tabcolsep}{4pt}
\begin{tabular}{l@{\hspace{0.25cm}}c@{\hspace{0.25cm}}c@{\hspace{0.25cm}}c}
\toprule
\multicolumn{4}{c}{\bf English CoNLL 2003}\\\hline
{\bf Model} & {\bf Prec.} & {\bf Rec.} & {\bf F1}  \\
\midrule
ELMo\cite{peters2018deep} & - & - & 92.22   \\
CVT\cite{kevin2018cross} & - & - & 92.6 \\
BERT-Tagger\cite{devlin2018bert}& - & - & 92.8  \\
BERT-MRC\cite{xiaoya2019ner} & 92.33 & 94.61 & {\bf 93.04}  \\\hline
% BERT-MRC+WeightCE & 93.32 & 92.78 & 93.05 \\
%  &  &  & (+0.01) \\\hline
BERT-MRC+FL & 93.13 & 93.09 & 93.11  \\
 &  & & (+0.06)  \\\hline
BERT-MRC+DL & 93.22 & 93.12 & 93.17  \\
 &  &  & (+0.12)  \\\hline
BERT-MRC+DSC & {93.41} & {93.25} & \textbf{93.33}  \\
&  &  & \textbf{(+0.29)}  \\
\bottomrule
\multicolumn{4}{c}{\bf English OntoNotes 5.0}\\\hline
{\bf Model} & {\bf Prec.} & {\bf Rec.} & {\bf F1} \\
\midrule
CVT \cite{kevin2018cross} & -& - & 88.8 \\
BERT-Tagger \cite{devlin2018bert} & 90.01& 88.35& 89.16 \\
BERT-MRC\cite{xiaoya2019ner} & 92.98 & 89.95 & {\bf 91.11}  \\\hline
% BERT-MRC+WeightCE & 89.99 & 92.92 & 91.43 \\
%  &  &  & (+0.32) \\\hline
BERT-MRC+FL & 90.13 & 92.34 & 91.22 \\
 &  &  & (+0.11)  \\\hline 
BERT-MRC+DL  & 91.70 & 92.06 & 91.88  \\
  &  &  & (+0.77)  \\\hline
BERT-MRC+DSC & {91.59} & {92.56} & \textbf{92.07}   \\
 &  &  & \textbf{(+0.96)}   \\
\bottomrule
\multicolumn{4}{c}{\bf Chinese MSRA}\\\hline
{\bf Model} & {\bf Prec.} & {\bf Rec.} & {\bf F1}  \\
\midrule
Lattice-LSTM \cite{lattice2018zhang}  & 93.57& 92.79 & 93.18  \\
BERT-Tagger \cite{devlin2018bert} & 94.97 & 94.62 & 94.80 \\
Glyce-BERT \cite{wu2019glyce}& 95.57 & 95.51 & 95.54\\
BERT-MRC\cite{xiaoya2019ner} & 96.18 & 95.12 & {\bf 95.75}  \\\hline
% BERT-MRC+WeightCE & 96.08 & 94.79 & 95.43 \\
%  &  &  & (-0.32) \\\hline
BERT-MRC+FL & 95.45 & 95.89 & 95.67  \\
 &  &  & (-0.08)  \\\hline
BERT-MRC+DL & 96.20 & 96.68 & 96.44  \\
 &  & & (+0.69)  \\\hline
BERT-MRC+DSC   & 96.67 & {96.77} & \textbf{96.72}  \\
  &  &  & \textbf{(+0.97)}  \\
\bottomrule

\multicolumn{4}{c}{\bf Chinese OntoNotes 4.0}\\\hline
{\bf Model} & {\bf Prec.} & {\bf Rec.} & {\bf F1}  \\
\midrule
Lattice-LSTM \cite{lattice2018zhang}  & 76.35& 71.56& 73.88  \\
BERT-Tagger \cite{devlin2018bert} & 78.01& 80.35& 79.16 \\
Glyce-BERT \cite{wu2019glyce}& 81.87 & 81.40 & 80.62 \\
BERT-MRC\cite{xiaoya2019ner} & 82.98 & 81.25 & {\bf 82.11}  \\\hline
% BERT-MRC+WeightCE & 83.45 & 83.87 & 83.66 \\
%  &  &  & (+1.55) \\\hline
BERT-MRC+FL & 83.63 & 82.97 & 83.30  \\
 &  &  & (+1.19)  \\\hline
BERT-MRC+DL & 83.97 & 84.05 & 84.01 \\
 &  &  & (+1.90) \\\hline
BERT-MRC+DSC   & {84.22} & {84.72} & \textbf{84.47}  \\
  &  &  & \textbf{(+2.36)}  \\
\bottomrule
\end{tabular}
}

\caption{Experimental results for NER task.}
\label{tab:ner_result}
\vskip -0.15in
\end{table}

To address this issue, we propose to multiply the soft probability $p$ with
 a decaying factor $(1-p)$, changing Eq.\ref{soft} to the following adaptive variant of DSC:
 \begin{equation}
 \text{DSC}(x_i) = \frac{2(1-p_{i1})^{\alpha} p_{i1}\cdot y_{i1}+\gamma}{(1-p_{i1})^{\alpha} p_{i1}+y_{i1}+\gamma}
 \label{adjust}
 \end{equation}
One can think $(1-p_{i1})^{\alpha}$ as a weight associated with each example, which changes as training proceeds. 
The intuition of changing $p_{i1}$ to $(1-p_{i1})^{\alpha} p_{i1}$ is to push down the weight of easy examples.
For
easy examples
whose probability are approaching
 0 or 1, $(1-p_{i1})^{\alpha} p_{i1}$  makes the model attach significantly less focus to them.

 A close look at Eq.\ref{adjust} reveals that it  actually mimics  the idea of focal loss (FL for short) \cite{lin2017focal} for object detection in vision. 
Focal loss was proposed for one-stage object detector 
to handle
foreground-background tradeoff encountered
during training.
It 
 down-weights the
loss assigned to well-classified examples by adding a $(1-p)^{\alpha}$ factor, leading the final loss to be $-(1-p)^{\alpha}\log p$.

\section{Experiments}
We evaluated the proposed method on four NLP tasks,  part-of-speech tagging, named entity recognition, machine reading comprehension and paraphrase identification. 
Hyperparameters are tuned on the corresponding development set of each dataset. 
More experiment details including datasets and hyperparameters are shown in supplementary material.

\begin{table*}[t]
\centering
\small
\begin{tabular}{lcccccc}
\toprule
 & \multicolumn{2}{c}{\bf SQuAD v1.1}& \multicolumn{2}{c}{\bf SQuAD v2.0}& \multicolumn{2}{c}{\bf QuoRef}  \\
{\bf Model}& {\bf EM} & {\bf  F1} &{\bf  EM} &{\bf   F1} & {\bf EM} &{\bf  F1} \\
\midrule
QANet \cite{qanet2018} & 73.6 & 82.7 & - & -&34.41 &38.26 \\
BERT \cite{devlin2018bert} &  {\bf 84.1} &{\bf 90.9}  & {\bf 78.7} & {\bf 81.9}& {\bf 58.44}& {\bf 64.95}\\\hline 
BERT+FL & 84.67 & 91.25 & 78.92 & 82.20 &60.78 & 66.19  \\
 & (+0.57) &(+0.35) & (+0.22) & (+0.30)& (+2.34) & (+1.24)  \\\hline
BERT+DL  & 84.83 & 91.86  & 78.99 & 82.88& 62.03 & 66.88 \\
  & (+0.73) & (+0.96)  & (+0.29) & (+0.98)& (+3.59) & (+1.93) \\\hline
BERT+DSC  & \textbf{85.34} & \textbf{91.97}  & \textbf{79.02} & \textbf{82.95}& \textbf{62.44} & \textbf{67.52}\\
  & \textbf{(+1.24)} & \textbf{(+1.07)}  & \textbf{(+0.32)} & \textbf{(+1.05)}& \textbf{(+4.00)} & \textbf{(+2.57)}\\
 \bottomrule
  % \hline\hline
% & \multicolumn{2}{c}{\bf SQuAD v1.1}& %\multicolumn{2}{c}{\bf SQuAD v2.0}& \multicolumn{2}{c}{\bf QuoRef}  \\
%{\bf Model}& {\bf EM} & {\bf  F1} & {\bf EM} & {\bf  F1} & {\bf EM} &{\bf  F1} \\\hline
XLNet \cite{xlnet2019} & {\bf 88.95}& {\bf 94.52}& {\bf 86.12}& {\bf 88.79}&{\bf 64.52}& {\bf 71.49} \\\hline
XLNet+FL & 88.90 & 94.55 & 87.04 & 89.32& 65.19 & 72.34 \\
 & (-0.05) & (+0.03) & (+0.92) & (+0.53)& (+0.67) & (+0.85)  \\\hline
XLNet+DL  & 89.13 & 95.36  & 87.22 & 89.44& 65.77 & 72.85 \\
  & (+0.18) & (+0.84)  & (+1.10) & (+0.65)& (+1.25) & (+1.36) \\\hline
XLNet+DSC  & \textbf{89.79} & \textbf{95.77}& \textbf{87.65} & \textbf{89.51} & \textbf{65.98} & \textbf{72.90}\\
  & \textbf{(+0.84)} & \textbf{(+1.25)}& \textbf{(+1.53)} & \textbf{(+0.72)} & \textbf{(+1.46)} & \textbf{(+1.41)}\\

\bottomrule\hline

\end{tabular}

\caption{Experimental results for MRC task.}
\label{tab:mrc_result}
\vskip -0.15in
\end{table*}
\begin{table}[t]
\centering
\small
\begin{tabular}{lcc}
\toprule
& \multicolumn{1}{c}{\bf MRPC}& \multicolumn{1}{c}{\bf QQP}\\
{\bf Model} & {\bf F1} & {\bf F1} \\
\midrule
BERT \cite{devlin2018bert} &  {\bf 88.0} & {\bf  91.3}\\\hline
BERT+FL &  88.43 &  91.86 \\
 &  (+0.43) &  (+0.56) \\\hline
BERT+DL  &  88.71 &  91.92 \\
 &  (+0.71) &  (+0.62) \\\hline
BERT+DSC  &  \textbf{88.92} &  \textbf{92.11} \\
  &  \textbf{(+0.92)} &  \textbf{(+0.81)} \\\bottomrule
% & \multicolumn{1}{c}{\bf MRPC}& \multicolumn{1}{c}{\bf QQP}\\
% {\bf Model} & {\bf F1} & {\bf F1} \\\hline
XLNet \cite{xlnet2019} &  {\bf 89.2} &  {\bf 91.8}  \\\hline
XLNet+FL &  89.25 &  92.31  \\
 &  (+0.05) & (+0.51)  \\\hline
XLNet+DL  &  89.33 &  92.39 \\
  &  (+0.13) &  (+0.59) \\\hline
XLNet+DSC  &  \textbf{89.78} &  \textbf{92.60} \\
 &  \textbf{(+0.58)} &  \textbf{(+0.79)} \\
\bottomrule\hline

\end{tabular}
\caption{Experimental results for PI task. }
\label{tab:paraphrase_result}
\vskip -0.15in
\end{table} 

\subsection{Part-of-Speech Tagging}
\label{pos}
\paragraph{Settings}
Part-of-speech tagging (POS) is the task of assigning a part-of-speech label (e.g., noun, verb, adjective) to each word in a given text. 
In this paper, we choose BERT \cite{devlin2018bert} as the backbone and conduct experiments on three widely used Chinese POS datasets including Chinese Treebank \cite{xue_xia_chiou_palmer_2005} 5.0/6.0 and UD1.4 and 
English datasets including Wall Street Journal (WSJ) and the dataset proposed by \newcite{posdata}.
% Wall Street Journal (WSJ) portion of the Penn Treebank 
We report the span-level micro-averaged precision, recall and F1 for evaluation. 

\paragraph{Baselines} We used the following baselines:
\begin{tightitemize}
\item {\bf Joint-POS:} \newcite{shao2017character} jointly learns Chinese word segmentation and POS.
\item {\bf Lattice-LSTM:} \newcite{lattice2018zhang} constructs a word-character lattice network. 
\item {\bf Bert-Tagger:} \newcite{devlin2018bert} treats part-of-speech as a tagging task.
\end{tightitemize}

\paragraph{Results} 
Table \ref{tab:pos_result} presents the experimental results on Chinese datasets. As can be seen, the proposed DSC loss outperforms the best baseline results by a large margin, i.e., outperforming BERT-tagger by +1.86 in terms of F1 score on CTB5,  +1.80 on CTB6  and   +2.19 on UD1.4. As far as we know, we are achieving SOTA performances on the three datasets. 
%  Weighted cross entropy and 
Focal loss only obtains a little performance improvement on CTB5 and CTB6, and the dice loss obtains huge gain on CTB5 but not on CTB6, which indicates the three losses are not consistently robust in solving the data imbalance issue. 
 % The proposed DSC loss performs robustly on all the three datasets.

Table \ref{tab:eng_pos_result} presents the experimental results for English datasets. 

\subsection{Named Entity Recognition}
\label{ner}
\paragraph{Settings}
Named entity recognition (NER) is the task of detecting the span and semantic category of entities within a chunk of text. 
Our implementation uses the current state-of-the-art model  proposed by \newcite{xiaoya2019ner} as the backbone, and changes the MLE loss to DSC loss. 
%For English datasets, we use BERT$_\text{Large}$ English checkpoints, while for Chinese we use the official Chinese checkpoints. 
Datasets that we use include
 OntoNotes4.0 \cite{ontonotes4}, MSRA \cite{msra2006ner}, CoNLL2003 \cite{conll2003ner} and OntoNotes5.0 \cite{ontonotes5}.
We report span-level micro-averaged precision, recall and F1. 

\paragraph{Baselines} We use the following baselines:
\begin{tightitemize}
\item {\bf ELMo:} a tagging model with pretraining from \newcite{peters2018deep}. 
\item {\bf Lattice-LSTM:} \newcite{lattice2018zhang} constructs a word-character lattice, only used in Chinese datasets.
\item {\bf CVT:} \newcite{kevin2018cross} uses Cross-View Training(CVT) to improve the representations of a Bi-LSTM encoder. 
\item {\bf Bert-Tagger:} \newcite{devlin2018bert} treats NER as a tagging task. 
\item {\bf Glyce-BERT:} \newcite{wu2019glyce} combines Chinese glyph information with BERT pretraining. 
\item {\bf BERT-MRC: }
\newcite{xiaoya2019ner} formulates NER as a machine reading comprehension task and achieves SOTA results on Chinese and English NER benchmarks. 
% The current SOTA model for both Chinese and English NER datasets proposed by \newcite{xiaoya2019ner}, which formulates NER as a machine reading comprehension (MRC) task.  
\end{tightitemize}

\paragraph{Results} Table \ref{tab:ner_result} shows experimental results on NER datasets. 
DSC outperforms BERT-MRC\cite{xiaoya2019ner} by +0.29, +0.96, +0.97 and +2.36 respectively on CoNLL2003, OntoNotes5.0, MSRA and OntoNotes4.0. 
As far as we are concerned, we are setting new SOTA performances on all of the four NER datasets.

\subsection{Machine Reading Comprehension}
\label{sec:mrc}
\paragraph{Settings}
The task of 
machine reading comprehension (MRC) \cite{seo2016bidirectional,wang2016multi,wang2016machine,wang2016multi,shen2017reasonet,chen2017reading} predicts the answer span in the passage given a question and the passage. 
We followed the standard protocols in \newcite{seo2016bidirectional}, in which the start and end indexes of answer are predicted. 
We report Extract Match (EM) as well as F1 score on validation set. 
We use three datasets on this task: SQuAD v1.1, SQuAD v2.0 \cite{rajpurkar2016squad, rajpurkar2018know} and Quoref \cite{dasigi2019quoref}.

\paragraph{Baselines} We used the following baselines: 
\begin{tightitemize}
\item {\bf QANet:} \newcite{qanet2018} builds a model based on convolutions and self-attentions.  Convolutions are used to model local interactions and self-attention
are used
to model global interactions. 
\item {\bf BERT:} \newcite{devlin2018bert} scores each candidate span and the maximum scoring span is used as a prediction.
\item {\bf XLNet:} \newcite{xlnet2019} proposes a generalized autoregressive pretraining method that enables learning bidirectional contexts. 
\end{tightitemize}

\paragraph{Results} Table \ref{tab:mrc_result} shows the experimental results for MRC task. With either BERT or XLNet, our proposed DSC loss obtains significant performance boost on both EM and F1.
For SQuADv1.1, our proposed method outperforms XLNet by +1.25 in terms of F1 score and +0.84 in terms of EM.
For SQuAD v2.0, the proposed method
achieves 87.65 on EM and 89.51 on F1. On QuoRef, the proposed method surpasses XLNet  by +1.46 on EM and +1.41 on F1. 
% by an average score above 1.0 in terms of both EM and F1, which indicates the DSC loss is complementary to the model structures.

\subsection{Paraphrase Identification}
\paragraph{Settings}
% Paraphrases are textual expressions that have the same semantic meaning using different surface words. 
Paraphrase identification (PI) is the task of identifying whether two sentences have the same meaning or not. 
We conduct experiments on the two widely-used datasets: MRPC \cite{mrpc2005} and QQP. 
 F1 score is reported for comparison.
We use BERT \cite{devlin2018bert} and XLNet \cite{xlnet2019} as  baselines. 

\paragraph{Results} Table \ref{tab:paraphrase_result} shows the results. 
We find that replacing the training objective with DSC introduces performance boost for both settings, +0.58 for MRPC and +0.73 for QQP.  

\section{Ablation Studies}
\begin{table*}[t]
\small
\centering
\begin{tabular}{lccccc}
\toprule
& \multicolumn{1}{c}{\bf original}& \multicolumn{1}{c}{\bf + positive}& \multicolumn{1}{c}{\bf + negative}& \multicolumn{1}{c}{\bf - negative}& \multicolumn{1}{c}{\bf + positive \& negative} \\ %\hline
% Model &  F1 & F1 &  F1  \\
\midrule
{ BERT} & 91.3 & 92.27 & 90.08 & 89.73 & 93.14\\\hline
{ BERT+FL} &  91.86(+0.56) & 92.64(+0.37) & 90.61(+0.53) & 90.79(+1.06) & 93.45(+0.31)  \\
{ BERT+DL} & 91.92(+0.62) & 92.87(+0.60) & 90.22(+0.14) & 90.49(+0.76) &  93.52(+0.38) \\
{ BERT+DSC} & 92.11(+0.81)  & 92.92(+0.65) & 90.78(+0.70) & 90.80(+1.07) & 93.63(+0.49)  \\
\bottomrule\hline
\end{tabular}
\caption{The effect of different data augmentation ways for QQP in terms of F1-score. }
%{\bf original training set} is short for \emph{original}; {\bf + positive} is short for \emph{add positive examples to balanced training set}; {\bf + negative} is short for \emph{add negative examples to more imbalanced training set}; {\bf - negative} is short for \emph{drop negative examples to balanced training set}; {\bf + positive \& negative} is short for {add positive and negative examples without changing data distribution.}}
\label{tab:ab_qqp}
\vskip -0.15in
\end{table*}
\subsection{Datasets  imbalanced to different extents}
% It is interesting to see how different data balancing techniques can be combined with the proposed training objectives to collaboratively address the data imbalance issue. 
% we introduce a dynamic weight adjusting strategy with dice loss to collaboratively address the data imbalance issue. 
It is interesting to see how differently the proposed objectives affect datasets  imbalanced to different extents.
We use the paraphrase identification dataset QQP (37\% positive and 63\% negative) for studies.
To construct datasets  with different imbalance degrees, we used the original QQP dataset to
construct synthetic training sets with different positive-negative ratios. 
Models are trained on these different synthetic sets and then test on the  same original test set.

\begin{tightitemize}
\item {\bf Original training set (original)}
The original dataset with
363,871 examples,  
with 37\%  being positive and 63\% being negative
\item {\bf Positive augmentation (+ positive)}\\
We created a balanced dataset by adding positive examples. 
We first randomly chose positive training examples in the original training set as templates. 
Then we used Spacy\footnote{\url{https://github.com/explosion/spaCy}} to retrieve entity mentions and replace them with new ones by linking mentions to their corresponding entities in DBpedia. 
The augmented set contains 458,477 examples, with 50\%  being positive and 50\% being negative.

\item {\bf Negative augmentation  (+ negative)}\\
We created a more imbalanced dataset. 
The size of 
the newly constructed 
training set and the data augmented technique are exactly the same as {\bf +negative}, 
except that we chose negative training examples as templates. 
The
augmented training set contains 458,477 examples, with 21\%  being positive and 79\% being negative.

\item {\bf  Negative downsampling  (- negative)}\\
We down-sampled negative examples in the original training set to get a balanced training set.
The down-sampled set contains 269,165 examples, with 50\%  being positive and 50\% being negative.

% The number of positive and negative examples is the same with the number of positive examples in {\it positive examples augmentation}.
% negative samples to balance we remove some negative samples from train set.
%\item {\bf Add positive and negative examples without changing data distribution  (+ positive /& negative)}\\
\item {\bf Positive and negative augmentation  (+ positive \& +negative)}\\
We augmented the original training data with additional positive and negative examples with the data distribution staying the same.
The augmented dataset contains 458,477 examples, with 50\%  being positive and 50\% being negative.

\end{tightitemize}

% The label distribution can introduce huge impact to the training process. 
% In this subsection, we investigate the impact using different data distribution. 
Results are shown in Table \ref{tab:ab_qqp}. 
We first look at the 
 first line, with all results  obtained using the MLE objective. 
 We can see that \textbf{+ positive} outperforms \textbf{original}, and \textbf{+negative} underperforms \textbf{original}.
 This is in line with our expectation since 
\textbf{+ positive} creates a balanced dataset while \textbf{+negative} creates a more imbalanced dataset.
Despite the fact  that \textbf{-negative} creates a balanced dataset, the number of training data decreases, resulting in inferior performances. 

DSC achieves the highest F1 score across all datasets.
Specially,
for {\bf +positive},  DSC achieves minor improvements (+0.05 F1) over  DL. 
In contrast,  it significantly outperforms  DL  for {\bf +negative} dataset. 
This is in line with our expectation since DSC  helps more on more imbalanced datasets. 
The performance of FL and DL are not consistent across different datasets, while DSC consistently performs the best on all datasets. 

\subsection{Dice loss for accuracy-oriented tasks?}
We argue that the cross-entropy objective is actually accuracy-oriented, whereas the proposed losses perform as a soft version of F1 score. To explore the effect of the dice loss on accuracy-oriented tasks such as text classification, we conduct experiments on the Stanford Sentiment Treebank (SST) datasets including SST-2 and SST-5. We fine-tuned BERT$_\text{Large}$  with different training objectives.  
\begin{table}[t]
\centering
\small
\begin{tabular}{lcc}
\toprule
 & \multicolumn{1}{c}{\bf SST-2}& \multicolumn{1}{c}{\bf SST-5} \\\hline
{\bf Model} &  Acc & Acc \\
\midrule
{ BERT+CE} & \textbf{94.90} & {\bf 55.57} \\
{ BERT+DL} & 94.37 & 54.63  \\
{ BERT+DSC} & 94.84  & 55.19   \\
\bottomrule\hline
\end{tabular}
\caption{The effect of DL and DSC on sentiment classification tasks.
BERT+CE refers to fine-tuning BERT and setting cross-entropy as the training objective. } 
\label{tab:loss_sst}
\vskip -0.15in
\end{table}
Experimental results for SST are shown in Table~\ref{tab:loss_sst}. For SST-5, BERT with CE achieves 55.57 in terms of accuracy, while DL and DSC perform slightly worse
(54.63 and 55.19, respectively).  
Similar phenomenon is observed for SST-2. 
These results verify that the proposed dice loss is not accuracy-oriented, and should not be used for accuracy-oriented tasks. 

\subsection{Hyper-parameters in Tversky Index}
% \subsection{The Effect of Tversky Index} 
%In this subsection, we examine the influence of the parameter at the interval of 0.1 in the objective L on the datasets. Experimetal results are shown in Figure \ref{fig:parameter}. Best performance is obtained when is equal to 0.1. 
As mentioned in Section \ref{sec33}, Tversky index (TI) offers the flexibility in controlling 
the tradeoff between false-negatives and false-positives. In this subsection, we explore the effect of hyperparameters (i.e., $\alpha$ and $\beta$) in TI to test how they manipulate the tradeoff.
We conduct experiments on the Chinese OntoNotes4.0 NER dataset and English QuoRef MRC dataset. 
Experimental results are shown in Table \ref{tab:tversky_index}. 
The highest F1 on Chinese OntoNotes4.0 is 84.67 when $\alpha$ is set to 0.6 while 
for QuoRef, the highest F1 is 68.44 when $\alpha$ is set to 0.4.  In addition, we can observe that the performance varies a lot as $\alpha$ changes in distinct datasets, which shows that the hyperparameters $\alpha,\beta$ acturally play an important role in TI. 
\begin{table}[t]
\begin{center}
\small
% \resizebox{\linewidth}{18mm}{
\begin{tabular}{ccccl}\toprule
{\bf $\alpha$}&{\bf Chinese Onto4.0}&{\bf English QuoRef}\\\hline
$\alpha=0.1$&  80.13& 63.23\\
$\alpha=0.2$& 81.17&  63.45\\
$\alpha=0.3$& 84.22&  65.88\\
$\alpha=0.4$& 84.52& \textbf{68.44} \\
$\alpha=0.5$& 84.47& 67.52 \\
$\alpha=0.6$& \textbf{84.67}&  66.35\\
$\alpha=0.7$& 81.81&  65.09\\
$\alpha=0.8$& 80.97&  64.13\\
$\alpha=0.9$& 80.21&  64.84\\\bottomrule
\end{tabular}
\end{center}
\caption{The effect of hyperparameters in Tversky Index. We set $\beta=1-\alpha$ and thus we only list $\alpha$ here.} 
\label{tab:tversky_index}
\vskip -0.15in
\end{table}

\section{Conclusion}
In this paper, we propose the dice-based loss to narrow down the gap between training objective  and evaluation metrics (F1 score). Experimental results  show that the proposed loss function help to achieve significant performance boost without changing model architectures. 

\section*{Acknowledgement}
We thank all anonymous reviewers, as well as Qinghong Han, Wei Wu and Jiawei Wu for their comments and suggestions. 
The work is supported by the National Natural Science Foundation of China (NSFC No. 61625107 and 61751209). 

% differentiable evaluation metric such as BLEU, ROUGE-L

% The proposed training objective leads to significant performance boost for part-of-speech, named entity recognition, machine reading comprehension and paraphrase identification tasks.  

% In this paper, we alleviate the severe data imbalance issue in NLP tasks. 
% We propose to use dice loss and its variants in replacement of the standard cross-entropy loss, which perform as a soft version of F1 score. Using dice loss can help  narrow the gap between training objectives and evaluation metrics. 
% Empirically, we show that the proposed training objective leads to significant performance boost for part-of-speech, named entity recognition, machine reading comprehension and paraphrase identification tasks. 
% We also explore the effect of the proposed losses on accuracy-oriented tasks such as text classification, and we find using the dice loss that performs as a soft version of F1 score, does harm the performance in terms of accuracy. 
% In future work, we'll study more general methods to alleviate the data imbalance issue regardless of evaluation metrics.


\bibliography{loss_bitex}
\bibliographystyle{acl_natbib}

\appendix 
\section{Dataset Details}
\label{app:detail}
\subsection{Part-of-Speech Tagging}
\paragraph{Datasets} We conduct experiments on three widely used benchmark, i.e., Chinese Treebank 5.0\footnote{\url{https://catalog.ldc.upenn.edu/LDC2005T01}}/6.0\footnote{\url{https://catalog.ldc.upenn.edu/LDC2007T36}} and UD1.4\footnote{\url{https://universaldependencies.org/}}.
\begin{itemize}
\item {\bf CTB5} is a Chinese dataset for tagging and parsing, which contains 507,222 words, 824,983 characters and 18,782 sentences extracted from newswire sources, including 698 articles from Xinhua (1994-1998), 55 articles from Information Services Department of HKSAR (1997) and 132 articles from Sinorama Magazine (1996-1998 \& 2000-2001).
\item {\bf CTB6} is an extension of CTB5, containing 781,351 words, 1,285,149 characters and 28,295 sentences. 
\item {\bf UD} is the abbreviation of Universal Dependencies, which is a framework for consistent annotation of grammar (parts of speech, morphological features, and syntactic dependencies) across different human languages. In this work, we use UD1.4 for Chinese POS tagging.
\end{itemize}

\subsection{Named Entity Recognition}

\paragraph{Datasets} For the NER task, we consider both Chinese datasets, i.e., OntoNotes4.0\footnote{\url{https://catalog.ldc.upenn.edu/LDC2011T03}} and MSRA\footnote{\url{http://sighan.cs.uchicago.edu/bakeoff2006/}} , and English datasets, i.e., CoNLL2003 \footnote{\url{https://www.clips.uantwerpen.be/conll2003/ner/}} and OntoNotes5.0\footnote{\url{https://catalog.ldc.upenn.edu/LDC2013T19}}.
\begin{itemize}
\item {\bf CoNLL2003} is an English dataset with 4 entity types: Location, Organization, Person and Miscellaneous. We followed data processing protocols in \cite{ma2016end}.
\item {\bf English OntoNotes5.0} consists of texts from a wide variety of sources and contains 18 entity types. We use the standard train/dev/test split of CoNLL2012 shared task.
\item {\bf Chinese MSRA} performs as a Chinese benchmark dataset containing 3 entity types. Data in MSRA is collected from news domain. Since the development set is not provided in the original MSRA dataset,  we randomly split the training set into training and development splits by 9:1. We use the official test set for evaluation.
\item {\bf Chinese OntoNotes4.0} is a Chinese dataset and consists of texts from news domain, which has 18 entity types. In this paper, we take the same data split as \newcite{wu2019glyce} did.
\end{itemize}

\subsection{Machine Reading Comprephension}

\paragraph{Datasets} For MRC task, we use three datasets:  SQuADv1.1/v2.0\footnote{\url{https://rajpurkar.github.io/SQuAD-explorer/}} and Queref\footnote{\url{https://allennlp.org/quoref}} datasets.
\begin{itemize}
\item {\bf SQuAD v1.1 and SQuAD v2.0} are the most widely used QA benchmarks. SQuAD1.1 is a collection of 100K crowdsourced question-answer pairs, and SQuAD2.0 extends SQuAD1.1 allowing  no short answer exists in the provided passage. 
\item {\bf Quoref} is a QA dataset which tests the coreferential reasoning capability of reading comprehension systems, containing 24K questions over 4.7K paragraphs from Wikipedia.
\end{itemize}

\subsection{Paraphrase Identification}

\paragraph{Datasets} Experiments are conducted on two PI datasets: MRPC\footnote{\url{https://www.microsoft.com/en-us/download/details.aspx?id=52398}} and QQP\footnote{\url{https://www.quora.com/q/quoradata/First-Quora-Dataset-Release-Question-Pairs}}.
\begin{itemize}
\item {\bf MRPC} is a corpus of sentence pairs automatically extracted from online news sources, with human annotations of whether the sentence pairs are semantically equivalent. 
The MRPC dataset has imbalanced classes (6800 pairs in total, and 68\% for positive, 32\% for negative).
\item {\bf QQP} is a collection of question pairs from the community question-answering website Quora. The class distribution in QQP is also unbalanced (over 400,000 question pairs in total, and 37\% for positive, 63\% for negative).
\end{itemize}

\end{document}